\documentclass{article}
\usepackage{microtype}
\usepackage{graphicx}
\usepackage{booktabs}
\usepackage{hyperref}

\usepackage[accepted]{icml2021}
\usepackage{multirow}
\usepackage{amsmath}
\usepackage{amssymb}
\usepackage{subcaption}

\newcommand{\figcrosstask}[3]{
\begin{figure}[tb]
    \centering
    \includegraphics[trim=#1, clip, width=#2\linewidth]{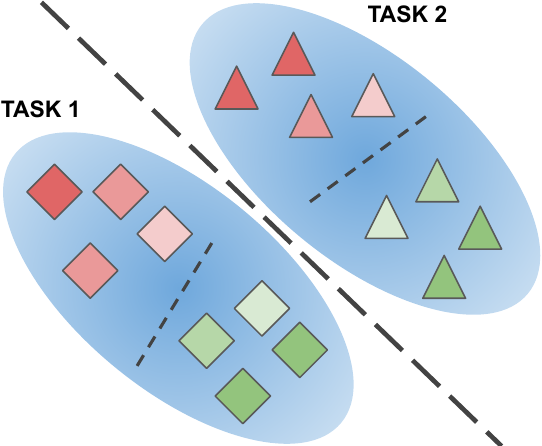}
    \caption{#3}
    \label{fig:cross_task}
    \vspace{-0.7em}
\end{figure}
}

\newcommand{\figcomparison}[3]{
\begin{figure}[tb]
\centering
    \includegraphics[trim=#1, clip, width=#2\linewidth]{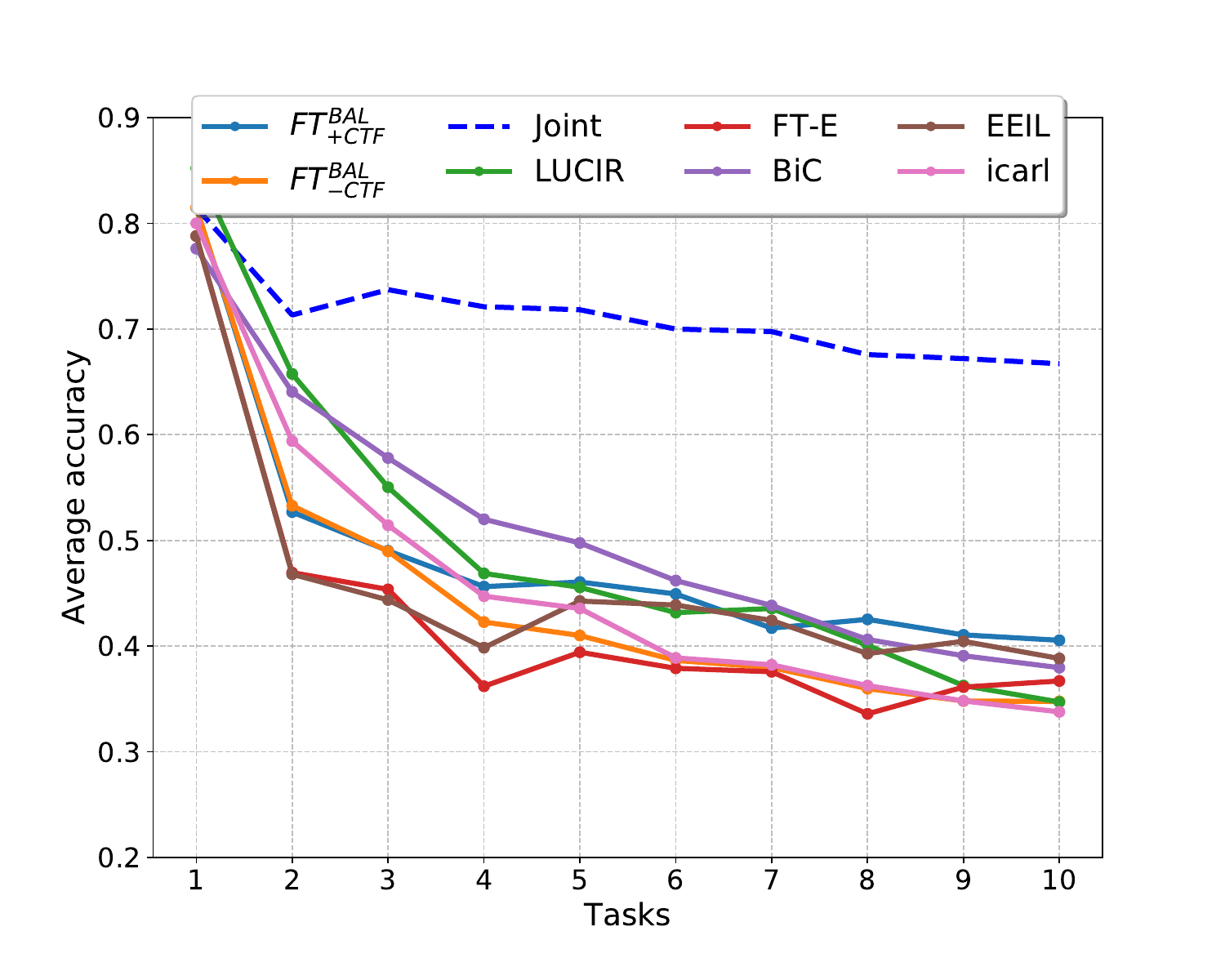}
    \vspace{-1.75em}
    \caption{#3}
    \label{fig:soa_comparison}
    \vspace{-0.25em}
\end{figure}
}

\newcommand{\figaccjoint}[3]{
\begin{figure*}[tb]
    \centering
    \includegraphics[trim=#1, clip, width=#2\linewidth]{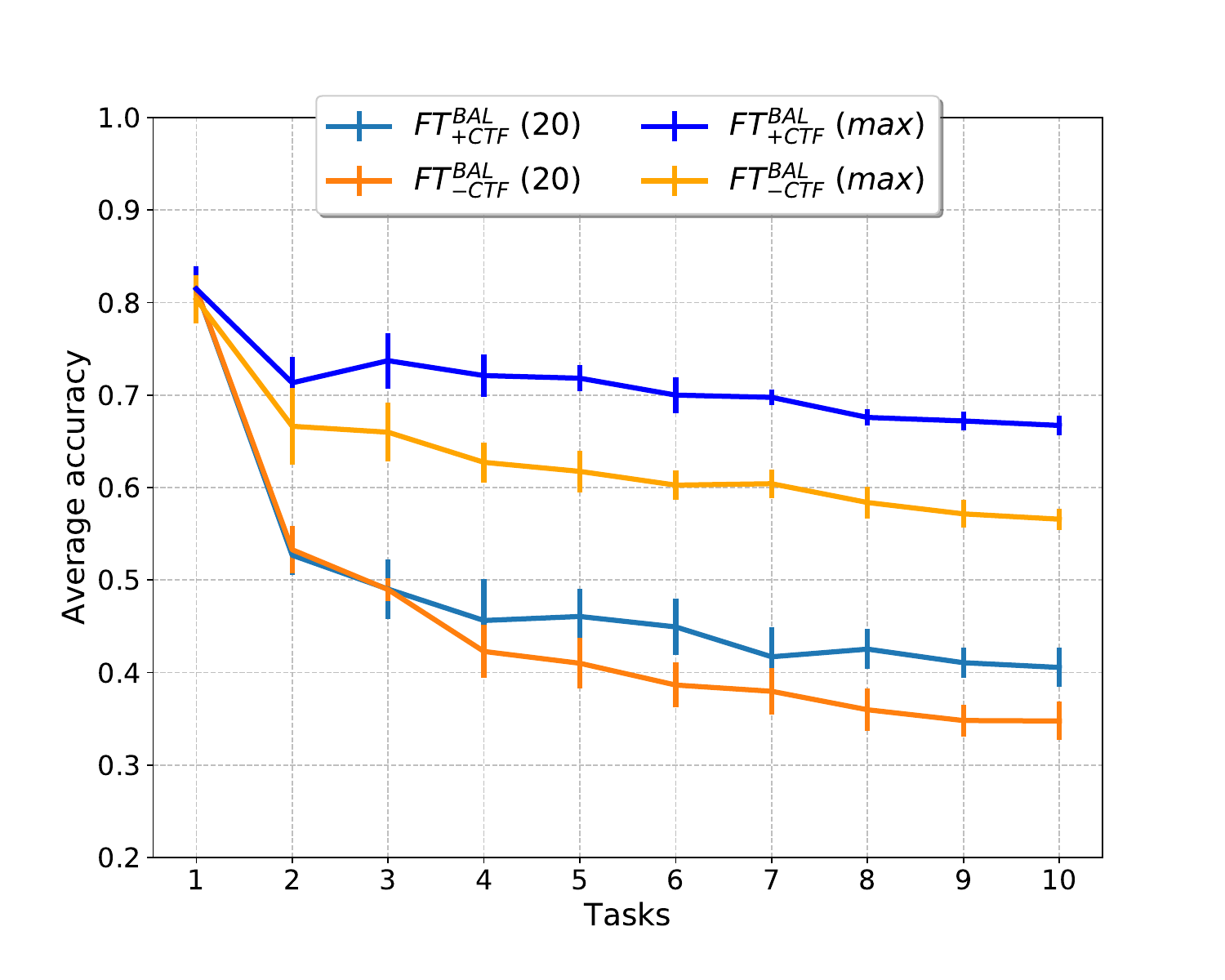}%
    \includegraphics[trim=#1, clip, width=#2\linewidth]{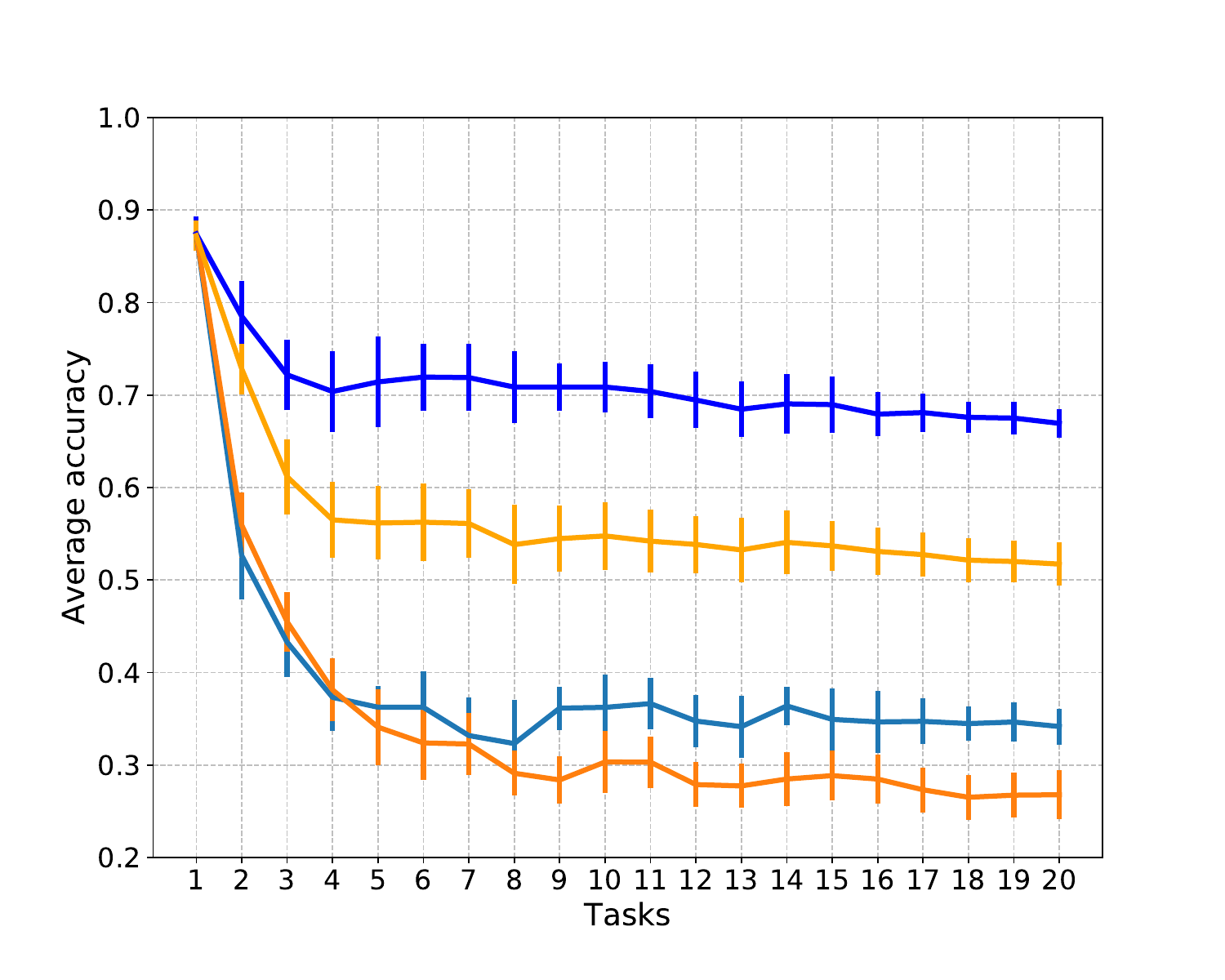}
    \vspace{-0.75em}
    \caption{#3}
    \label{fig:acc_joint}
\end{figure*}
}

\newcommand{\figtaskinf}[3]{
\begin{figure}[ht]
    \centering
    \includegraphics[trim=#1, clip, width=#2\linewidth]{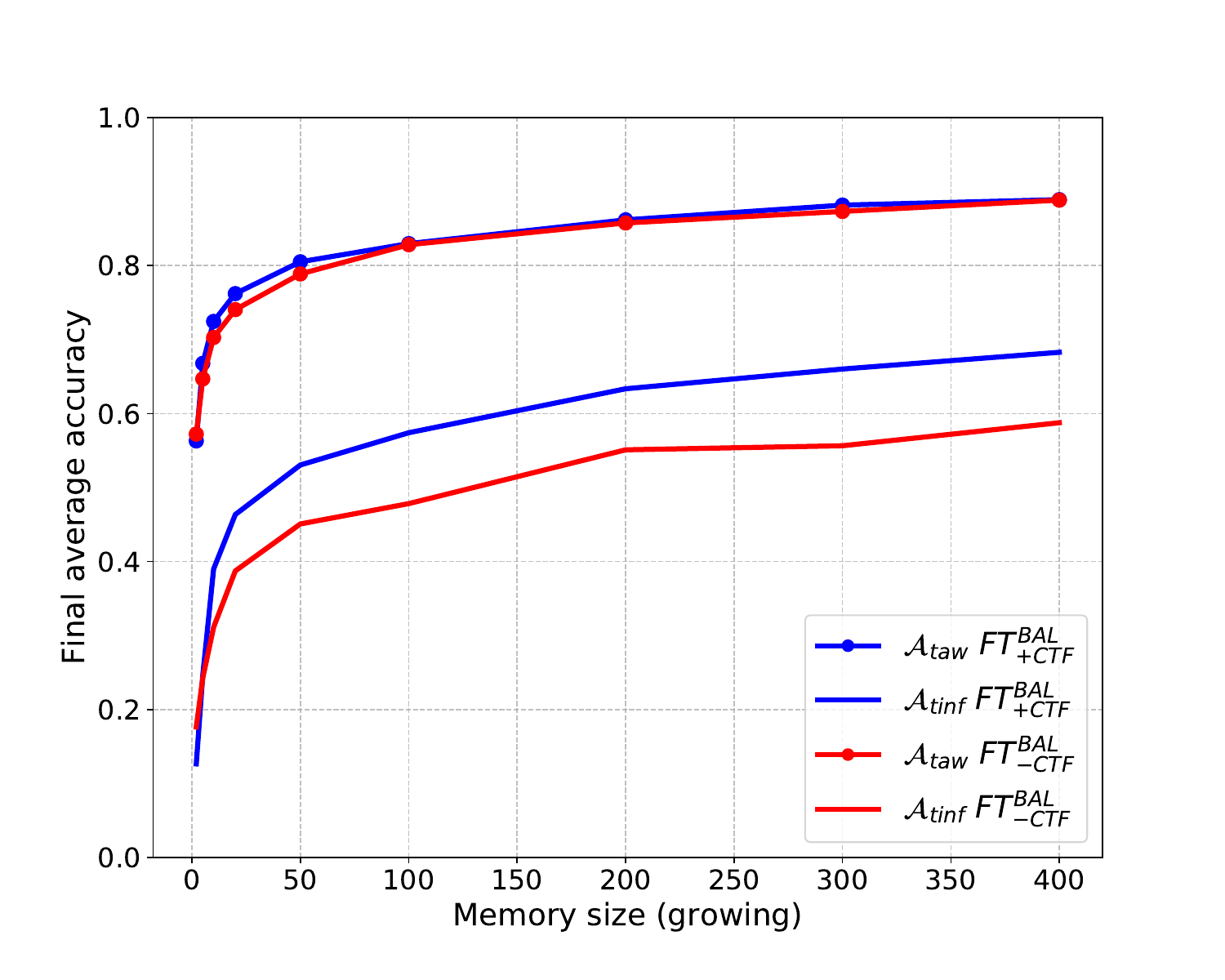}
    \caption{#3}
    \label{fig:task_inf}
    \vspace{-0.5em}
\end{figure}
}

\newcommand{\figimnet}[3]{
\begin{figure}[tb]
    \centering
    \includegraphics[trim=#1, clip, width=#2\linewidth]{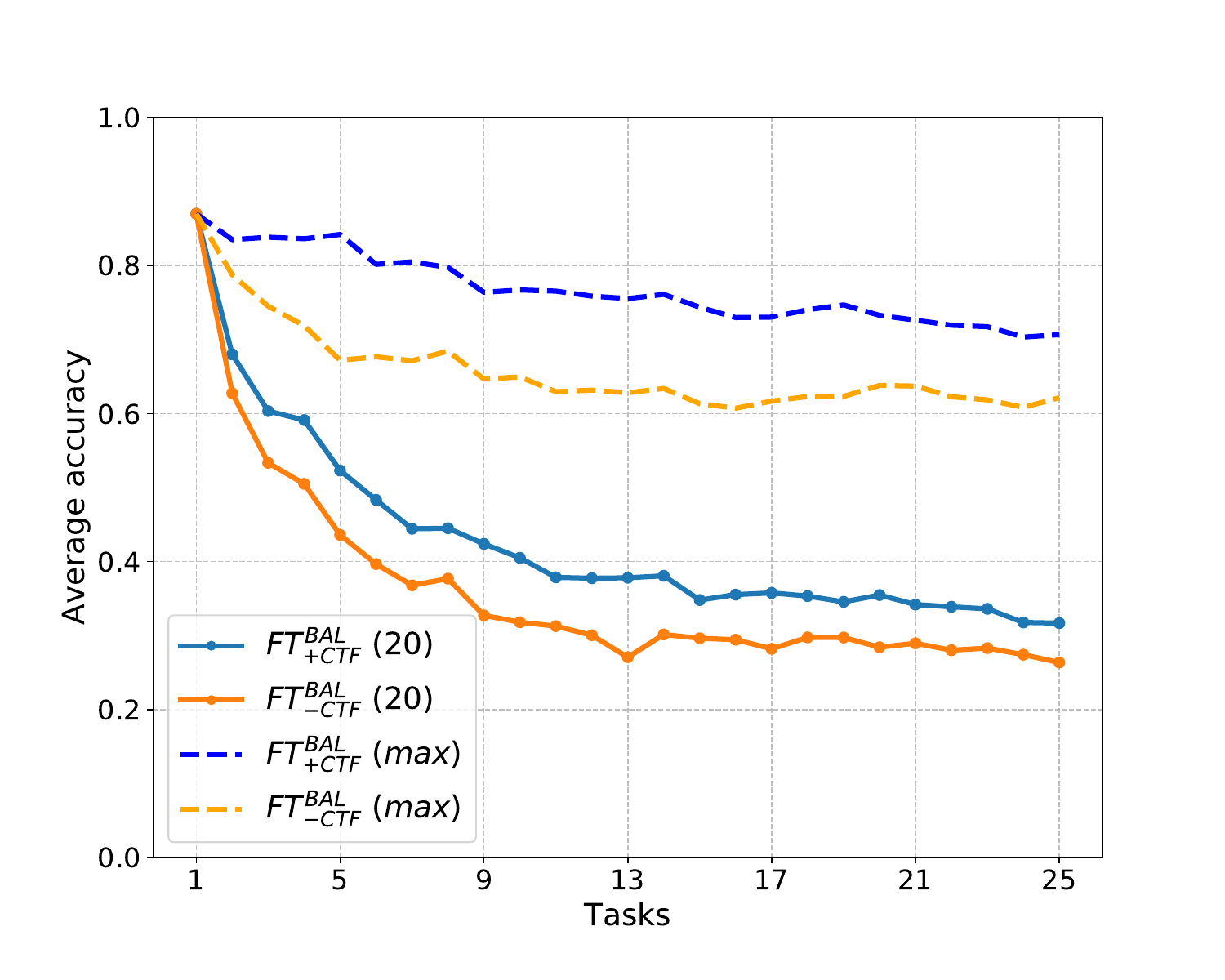}
    \vspace{-1.75em}
    \caption{#3}
    \label{fig:imagenet}
    \vspace{-0.5em}
\end{figure}
}

\newcommand{\figaccprognew}[4]{
\begin{figure}[tb]
    \centering
    \includegraphics[trim=#1, clip, width=#2\linewidth]{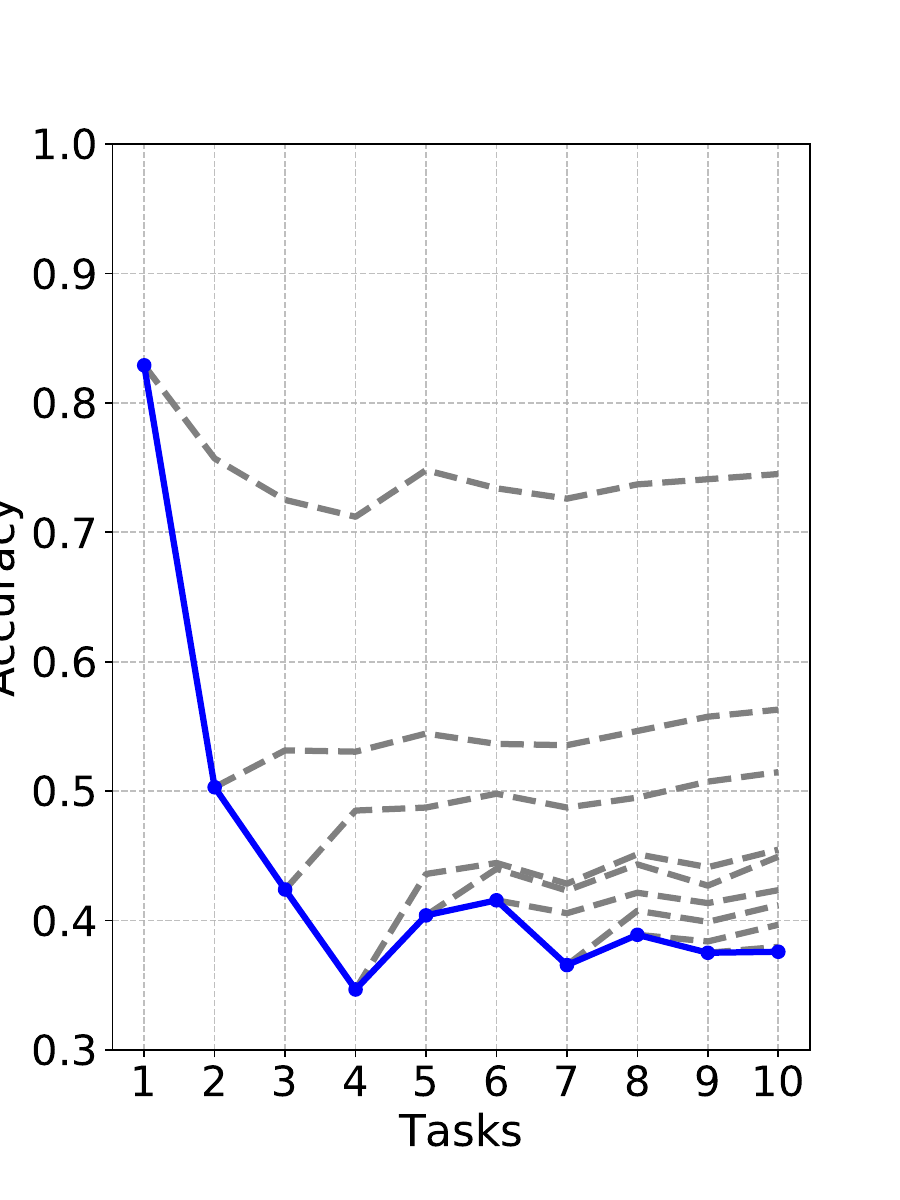}
    ~\includegraphics[trim=#3, clip, width=#2\linewidth]{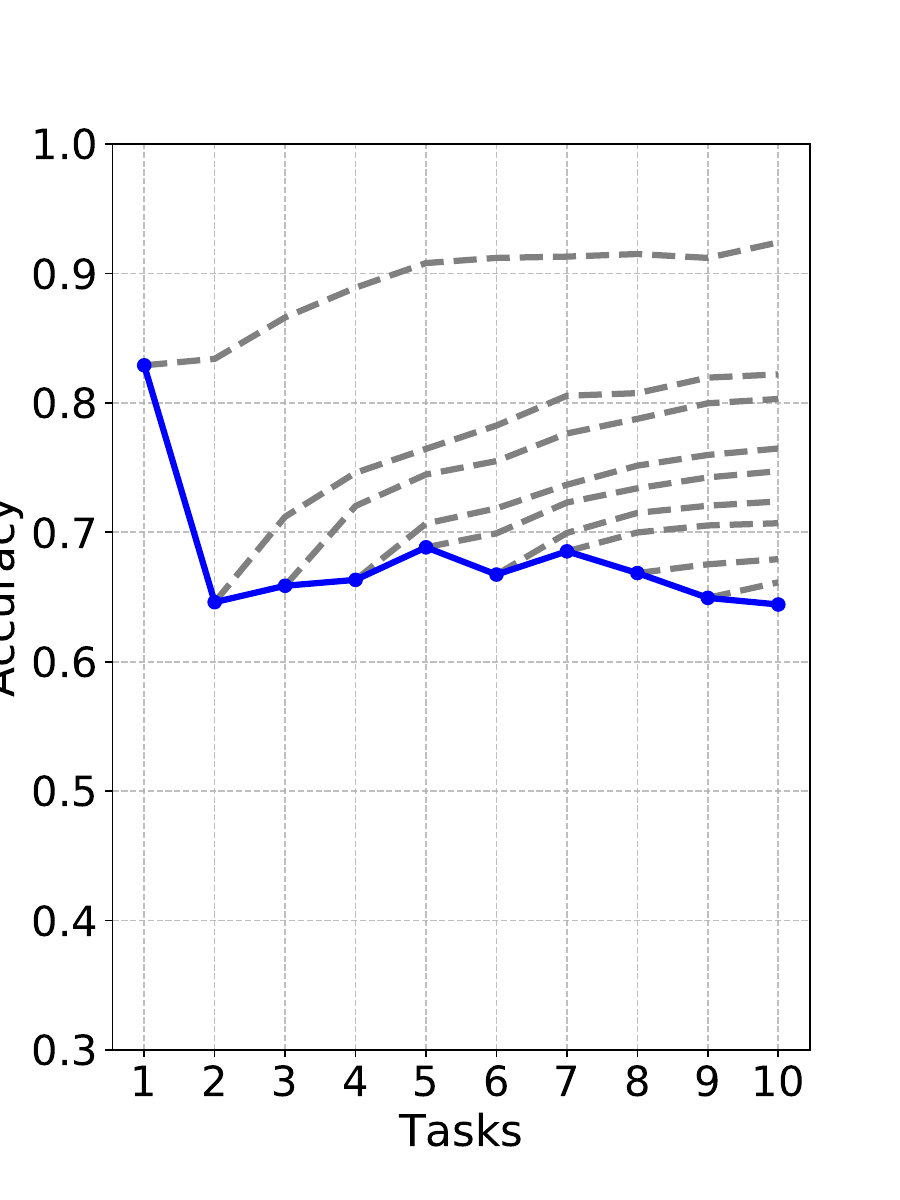}
    \caption{#4}
    \label{fig:cumulative_accs}
\end{figure}
}

\newcommand{\figgaps}[3]{
\begin{figure}[tb]
    \centering
    \includegraphics[trim=#1, clip, width=#2\linewidth]{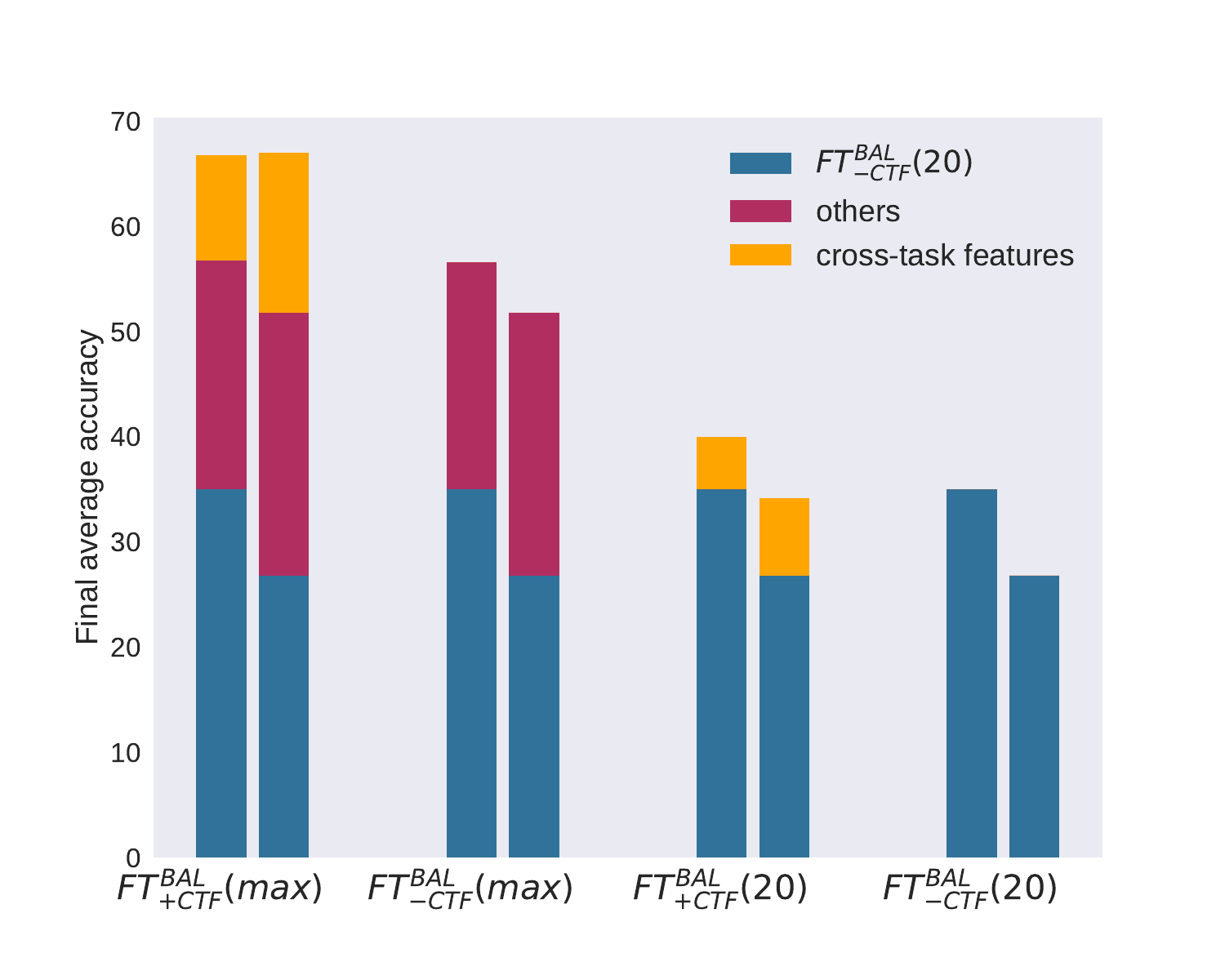}
    \caption{#3}
    \label{fig:gaps}
\end{figure}
}

\newcommand{\algorithmbalanced}[1]{
\begin{algorithm}[tb]
   \caption{#1}
   \label{procedure}
\begin{algorithmic}
   \STATE {\bfseries Input:} task sequence $\mathcal{T}$, data $D$, mem. buffer $\mathcal{M}$, loss $\mathcal{L}$
   \FOR{$t\in 1\dots T$}
   \FOR{$i \in 1 \dots N_{\text{epochs}}$}
   \FOR{$\mathcal{B} \sim D^t \cup \mathcal{M}$}
   \STATE $\theta' \gets \text{SGD}(\mathcal{L}, \mathcal{B}, \theta)$
   \ENDFOR
   \ENDFOR
   \STATE $\mathcal{M} \gets \text{FillMemory}(D^t, \mathcal{M})$
   \FOR{$i \in 1 \dots N_{\text{FTepochs}}$}
   \FOR{$\mathcal{B}_{mem} \sim \mathcal{M}$}
   \STATE $\theta_{\nu}'' \gets \text{SGD}(\mathcal{L}_{CE}, \mathcal{B}_{mem}, \theta')$
   \ENDFOR
   \ENDFOR
   \ENDFOR
\end{algorithmic}
\end{algorithm}
}

\newcommand{\tablecifargrowmore}[1]{
\begin{table*}
\centering
\caption{#1}
\vspace{-0.5em}
\begin{tabular}{c@{\hspace{0.7cm}}c@{\hspace{0.7cm}}ccccc}
\toprule
\multirow{2}{*}{scenario} & \multirow{2}{*}{approach} & \multicolumn{4}{c}{growing memory size} \\
 &  & 20/cls & 10/cls & 5/cls & 2/cls & max\\
\midrule
\multirow{2}{*}{10 tasks}
 & $FT^{BAL}_{\text{+}CTF}$           & 40.55 $\pm$ 2.1 & 31.5 $\pm$ 3.0 & 19.1 $\pm$ 2.5 &  9.0 $\pm$ 0.69 & 66.8 $\pm$ 1.1 \\
 & $FT^{BAL}_{\text{-}CTF}$ & 34.8 $\pm$ 2.1 & 27.4 $\pm$ 1.5 & 20.1 $\pm$ 1.2 & 14.3 $\pm$ 0.5 & 56.6 $\pm$ 1.2 \\
\midrule
\multirow{2}{*}{20 tasks}
 & $FT^{BAL}_{\text{+}CTF}$           & 34.2 $\pm$ 1.9 & 26.6 $\pm$ 2.7 & 12.5 $\pm$ 1.6 &  4.4 $\pm$ 0.15 & 67.0 $\pm$ 1.5 \\
 & $FT^{BAL}_{\text{-}CTF}$ & 26.8 $\pm$ 2.6 & 20.1 $\pm$ 2.0 & 15.1 $\pm$ 1.7 & 9.2 $\pm$ 0.7 & 51.8 $\pm$ 2.3 \\
\bottomrule
\end{tabular}
\label{tab:cifar100}
\end{table*}
}

\newcommand{\tableforgetting}[1]{
\begin{table}[tb]
\centering
\caption{#1}
\vspace{-0.75em}
\label{tab:forgetting}
\setlength\tabcolsep{4pt}
\resizebox{\linewidth}{!}{
\begin{tabular}{c@{\hspace{0.5cm}}cc@{\hspace{0.5cm}}cc@{\hspace{0.8cm}}cc@{\hspace{0.5cm}}cc}
\toprule
 & \multicolumn{2}{c}{10 Tasks} & \multicolumn{2}{c}{20 Tasks} \\
memory & \multicolumn{1}{c}{classic} & \multicolumn{1}{c}{cumulative} & \multicolumn{1}{c}{classic} & \multicolumn{1}{c}{cumulative}\\
size &  &  &  &  \\
\toprule
(2/class)   & 81.6
            & 4.0 
            & 86.0
            & 6.4 \\
\midrule
(5/class)   & 69.5
            & 1.6
            & 74.1
            & 1.1 \\
\midrule
(10/class)   & 56.0
             & 0.9
             & 49.5
             & 0.7 \\
\midrule
(20/class)
              & 28.8
              & 0.4
              & 39.9
              & 0.6 \\
\bottomrule
\end{tabular}}
\end{table}
}

\newcommand{\minisection}[1]{\vspace{0.04in} \noindent {\bf #1}\ \ }
\DeclareMathOperator*{\argmax}{arg\,max}

\icmltitlerunning{On the importance of cross-task features for class-incremental learning}

\begin{document}

\twocolumn[
\icmltitle{On the importance of cross-task features for class-incremental learning}
\icmlsetsymbol{equal}{*}
\begin{icmlauthorlist}
\icmlauthor{Albin Soutif--Cormerais}{cvc}
\icmlauthor{Marc Masana}{cvc,icg}
\icmlauthor{Joost van de Weijer}{cvc}
\icmlauthor{Bart{\l}omiej Twardowski}{cvc}
\end{icmlauthorlist}
\icmlaffiliation{cvc}{Computer Vision Center, Barcelona}
\icmlaffiliation{icg}{ICG, Graz University of Technology and DES Silicon Austria Labs}
\icmlcorrespondingauthor{Albin Soutif--Cormerais}{albin@cvc.uab.cat}
\icmlkeywords{Machine Learning, ICML, Lifelong Learning}

\vskip 0.3in
     ]
     
\printAffiliationsAndNotice{}

\begin{abstract}
In class-incremental learning, an agent with limited resources needs to learn a sequence of classification tasks, forming an ever growing classification problem, with the constraint of not being able to access data from previous tasks. The main difference with task-incremental learning, where a task-ID is available at inference time, is that the learner also needs to perform cross-task discrimination, i.e. distinguish between classes that have not been seen together. Approaches to tackle this problem are numerous and mostly make use of an external memory (buffer) of non-negligible size.
In this paper, we ablate the learning of cross-task features and study its influence on the performance of basic replay strategies used for class-IL.
We also define a new forgetting measure for class-incremental learning, and see that forgetting is not the principal cause of low performance. Our experimental results show that future algorithms for class-incremental learning should not only prevent forgetting, but also aim to improve the quality of the cross-task features, and the knowledge transfer between tasks. This is especially important when tasks contain limited amount of data.
\end{abstract}

\section{Introduction}
In order to allow agents with limited resources to update their knowledge in an efficient manner without access to previously learned data, we need to solve the problem of incremental learning. This setting assumes that an agent is exposed to a sequence of tasks which have to be learned one at a time while avoiding catastrophic forgetting of previous tasks~\cite{french1999catastrophic, icarl, thrun1996learning}. In practice, this is justified due to the inconvenience or infeasibility of storing data from each task and having to train a new model each time on all data seen so far, which requires a lot of computation and memory. Incremental Learning can be divided between task-incremental learning (task-IL)~\cite{delange2021continual} and class-incremental learning\footnote{In class-IL and task-IL, data comes as a sequence of batches of classes which we refer to as tasks. In this article, we do not consider the scenario in which these batches would contain only one class.} (class-IL)~\cite{class_incr_survey} depending on the availability or not of the task-ID at test time, respectively. In both cases, each task can be learned during the span of a training session.

\figcrosstask{0 0 0 0}{0.6}{Fictive scenario illustrating the two types of features considered. In this scenario, the color is an intra-task feature, and the shape is a cross-task feature. The intra-task features are insufficient to solve the final 4-class problem.}

A particular challenge of class-incremental learning is that classes in different tasks are never learned together but have to be discriminated from each other (no task-ID at test time). One way to tackle this is to store a small subset of instances of each class seen so far to later do rehearsal when learning new ones~\cite{end2end, icarl}. Therefore, the agent is capable of seeing classes from different tasks together over the course of training.
Even though some class-IL methods do not make use of a memory buffer~\cite{LwM, yu2020semantic}, these can currently not compete with methods using one~\cite{class_incr_survey}. Therefore, in this paper, we focus our attention on methods with a memory buffer.
However, because of the memory restrictions of such rehearsal memory buffer, a class imbalance is introduced during the learning process. One of the side effects of this class imbalance is that the biases and the norm of weights in the classifier tend to be higher in the classes from newer tasks~\cite{LUCIR, class_incr_survey, bic}. This phenomenon is called task-recency bias and is tackled with different bias-correction strategies by recent approaches~\cite{il2m, LUCIR, bic}. Having addressed the task-recency bias, there is still a large gap with respect to the upper-bound (joint training). In this paper, we investigate to what extend the training of better cross-task features can narrow this gap.

Another focus of this work is the study of the causes of performance drop in class-IL. As in other incremental learning settings, processing the sequence of tasks in class-IL leads to a dramatic drop in performance from earlier tasks to the latest ones. Many class-IL works associate \emph{catastrophic forgetting}, which has been widely studied in the task-IL setting, to this performance drop. However, the metrics defined for task-IL are not well suited to the class-IL setting and can give misleading results when applied directly to the latter. In this work, we will define similar metrics adapted to class-IL, and use them to increase understanding of the main causes of performance drop when moving to that setting.

When learning in an incremental manner, we consider two types of features that can be learned by the feature extractor (see Fig.~\ref{fig:cross_task}). A \emph{cross-task discriminative feature} discriminates between classes that belong to different tasks, while an \emph{intra-task discriminative feature} discriminates between classes from the same task. Furthermore, some features might appear during training that satisfy both types of discrimination. Given feature types, we postulate that replay in class-IL should fulfil multiple roles at the feature extractor level. First, to maintain previously learned intra-task discriminative features. Second, to create cross-task discriminative features capable of discriminating between classes not present in the same task. For instance, in Fig.~\ref{fig:cross_task}, discriminating between classes of the same task only requires to learn color features (intra-task), while solving the cumulative tasks proposed in class-IL requires to also learn shape features (cross-task). Its third role is to enable knowledge transfer between the new task and previous tasks so that one task can benefit from the learning of another as it occurs when learning multiple tasks concurrently \cite{multitask}. Failing to learn either type of feature would result in higher miss-classification rate. In this paper, we aim to study whether these two types of features are learned properly with replay, especially cross-task features.

Our contributions are summarised as follow:
\begin{itemize}
    \item We compare two baselines using replay. One aims to learn cross-task features while the other does not. This helps us to assess what is the importance of cross-task features in the class-IL setting.
    \item We question the use of classic task-IL metrics in \mbox{class-IL} and propose new appropriate metrics. We observe that \emph{catastrophic forgetting} is not the main cause of performance drop in class-IL.
\end{itemize}
In addition, we perform experiments allowing an increasing amount of memory to the class-incremental algorithms. These results suggest that even though there is no forgetting, there is room for improvements in task specific accuracy that can significantly improve cross-task discrimination without explicitly building features for it.

\section{Related Work}
Class-incremental learning approaches can be divided in three categories~\cite{class_incr_survey}: regularisation-based, bias-correction, and rehearsal-based, which are often combined to tackle multiple incremental learning challenges.

Regularisation-based approaches add a regularisation term to the loss which penalises changes in the weights~\cite{mas, rwalk, ewc} or activations~\cite{lfl, lwf, EBLL, DMC}. This maintains the stability-plasticity trade-off between maintaining previous knowledge and learning new tasks. These approaches have been widely used in task-IL. Furthermore, they have also shown promising results in class-incremental learning, whether by pairing with rehearsal-based approaches~\cite{end2end, LUCIR, icarl, bic} or with other strategies such as attention~\cite{LwM}.

Rehearsal has become the most commonly used approach in class-incremental learning, focusing on images or feature replay of previous tasks while learning a new one. Rehearsal-based approaches can be further divided into different subcategories. Pseudo-rehearsal methods replay generated images~\cite{shin2017continual}, thus avoiding the privacy issue of storing raw data. They also have the advantage of reducing storage memory and generating it on-the-fly instead. In MerGan~\cite{mergan}, a conditional GAN is used to replay images, introducing the learning and storage of an advanced network instead of having to store exemplars. In Generative Feature Replay~\cite{gen_feat_rep}, a GAN is used to replay smaller size features in to the classification layer, making the GAN learning easier at the cost of fixing the feature extractor, limiting the power of future learning. Some pseudo-rehearsal methods generate images by directly optimising in the image space using a loss that aims to prevent forgetting~\cite{arm, mnemonics}, thus avoiding the use of an external generator. Classical rehearsal methods store exemplars and replay them along with the data from the task at hand. However, this introduces a data imbalance problem between the few exemplars available for previously learned classes and the large amount of data for the new ones. For that reason some approaches combine rehearsal with bias correction~\cite{il2m, LUCIR, bic}, which also tackles task-recency bias. In IL2M~\cite{il2m}, a dual-memory is proposed, storing both images and class-statistics, which are used to rectify the scores of past classes. Noticeably, most rehearsal approaches use a distillation loss term on the outputs of the model, while IL2M obtains similar results with a classic fine-tuning cross-entropy loss instead.

A more challenging setting is introduced with online learning, where no more than one pass is allowed on the current task data. Therefore, instead of incremental training sessions which can iterate over several epochs on the data, each sample is seen once unless stored in the reduced external exemplar memory. This setting was originally explored for task-IL, where the task-ID is known at test time~\cite{agem, gem}. In GEM~\cite{gem} and A-GEM~\cite{agem}, gradients are modified in order to avoid both increasing the loss on the exemplars and over-fitting on them. Eventually, it is observed in \emph{tiny episodic memories}~\cite{tiny}, that just replaying exemplars, even when only a few are available, efficiently prevents forgetting in the task-IL setting. Recently, some approaches have been applied to online class-IL~\cite{max_inter, gss, remind}. In MIR~\cite{max_inter}, controlled sampling is performed to automatically rehearse samples from tasks currently undergoing the most forgetting. REMIND~\cite{remind} compresses features from the exemplars using learned quantization modules, which later are replayed.

In recent surveys~\cite{ftbal, class_incr_survey}, inter-task confusion and bias correction are identified among the main challenges of class-IL. We argue that additional challenges are the ones of knowledge transfer between tasks, and the learning of cross-task features (in contrast with task-recency bias).
Finally, GDumb~\cite{gdumb} introduces a baseline which only uses exemplars from the buffer, obtaining comparable performance to the state-of-the-art in the online setting. The proposed method also works for class-IL since it only needs the exemplars present in the memory buffer, however, it achieves a lower performance. Similarly, our proposed baselines are also applicable to both online and offline class-IL.

\section{Intra- and Cross-task training}
\subsection{Notation}
We assume that the data comes in the form of the following task sequence:
\begin{equation}
    \mathcal{T} = \{(C^1, D^1), (C^2, D^2), ... (C^n, D^n)\},
\end{equation}
where $n$ is the number of tasks, $C^t$ is the set containing the classes of task $t$, with the constraint of non-overlapping classes between different tasks ($\forall \ 1 \leq i \leq n$, $i \neq j$, $C^i \cap C^j = \varnothing$). Let $D^t$ be the dataset for task $t$, containing the labelled images.
The learner is a neural network, function of its parameters $\theta$ and the input $x$, which we will denote by $f(x;\theta)$.
We further split this network into a feature extractor $\Psi(x;\theta_\Psi)$ with weights $\theta_\Psi$, and a classification layer $\nu(x;\theta_\nu)$ with weights $\theta_\nu$. The logits are obtained by applying the feature extractor first, then the classification layer, obtaining $f(x;\theta) = \nu(\Psi(x;\theta_\Psi);\theta_\nu)$. The upper-scripted $\theta^t$ denote the weights of the model after seeing task t. Additionally, we denote the current task as $T$, a batch of data as $\mathcal{B}$, and the upper-scripted subset containing data from task $t$ as $\mathcal{B}^t = \{ (x, y) \in \mathcal{B}^t, y \in C^t \}$. So that we can consider all classes and data that the network has seen so far, we define $C_{\Sigma}^{k} = \bigcup_{t = 1}^{k}{C^t}$ and $D_{\Sigma}^{k} = \bigcup_{t = 1}^{k}{D^t}$, where $\Sigma$ stands for cumulative.

\subsection{Two Replay Baselines}
In order to better dissect the replay mechanism, we propose two different baselines to address class-IL. The two methods differ only in the loss they apply to the feature extractor. The first one explicitly tries to learn both cross-task and intra-task discriminative features, while the second one avoids learning cross-task discriminative features, restricting itself to learn only intra-task ones. They are defined as:
\begin{equation}
    \mathcal{L}_{CE}(\mathcal{B}; \theta) = \sum_{(x, y) \in \mathcal{B}}{\!\!\!\!\!\! - \log\frac{\exp f(x;\theta)_y}{\sum_{c \in C_{\Sigma}^{T}}{\exp f(x;\theta)_c}}},
\label{eq:joint}
\end{equation}
\begin{equation}
    \mathcal{L}_{CE\text{-}IT}(\mathcal{B}; \theta) = \frac{1}{T}\sum_{t = 1}^{T}{\,\, \sum_{(x, y) \in \mathcal{B}^t}{\!\!\!\!\!\! - \log\frac{\exp f(x;\theta)_y}{\sum_{c \in C^{t}}{\exp f(x;\theta)_c}}}}.
\label{eq:multitask}
\end{equation}
$\mathcal{L}_{CE}$ (Eq.~\ref{eq:joint}) is the classical cross-entropy over all classes seen so far, which seems natural to use whenever using exemplars. $\mathcal{L}_{CE\text{-}IT}$ (Eq.~\ref{eq:multitask}) is the sum of the cross-entropies of each separate tasks, as it would be done in multi-task training or task-incremental learning, this loss is only learning intra-task discriminative features ($CE\text{-}IT$ stands for cross-entropy intra-task). During training, $\mathcal{B}$ will be drawn from $D^t \cup \mathcal{M}$ where $\mathcal{M}$ is a memory buffer that retains a small number of samples per class.

To tackle task-recency bias that might arise from unbalanced training, and calibrate the classification heads on top of the feature extractor, we add an extra step to our proposed baseline algorithms. Taking inspiration from  EEIL~\cite{end2end}, we perform an additional balanced fine-tuning step after learning the current task, training the network on data only from the exemplars memory buffer, which contains a balanced number of training samples per class. We use $\mathcal{L}_{CE}$ on both cases for this step, but only back-propagating it through the classifier, leaving the feature extractor frozen with the previously learned knowledge. This is a similar procedure as $FT^{BAL}$~\cite{ftbal} with the difference of the feature extractor parameters being frozen while the balancing step takes place. 

The detailed training procedure is explained in Algorithm~\ref{procedure}, where $\mathcal{L}$ is the loss used during the first training step, and is replaced by $\mathcal{L}_{CE}$ or $\mathcal{L}_{CE\text{-}IT}$ depending on the baseline we are using. We will later refer to these baselines as $FT^{BAL}_{\text{+}CTF}$ (using $\mathcal{L}_{CE}$) and $FT^{BAL}_{\text{-}CTF}$ (using $\mathcal{L}_{CE\text{-}IT}$) where $\text{+}CTF$ and $\text{-}CTF$ stand for with and without Cross Task Features respectively. Indeed, $FT^{BAL}_{\text{-}CTF}$ will emulate the learning of a feature extractor as it would be done in the task-incremental learning setting. This will allow us to evaluate how features learned similarly as in the task-IL setting perform when used in the class-IL scenario, without access to task id. This fine-tuning step can be seen as a sort of probing of the feature extractor, similarly to what is done in \cite{probes}, though here it is performed on a reduced dataset.

Naturally, we expect $FT^{BAL}_{\text{+}CTF}$ to perform better in the class-IL setting since it is targeted to that setting. Our interest will rather focus on the gap between $FT^{BAL}_{\text{+}CTF}$ and $FT^{BAL}_{\text{-}CTF}$. Studying this gap and its contribution to the larger gap between these methods and the upper bound for this setting will help us understand how  improvements in task-incremental learning can benefit class-incremental learning.

\algorithmbalanced{Two step procedure}

Finally, we shortly recall the main challenges for task-IL since these are inherited by class-IL. In task-IL, since the task-ID is available at test time, it is not necessary to learn any cross-task features to obtain good performance on each specific tasks. This setting stills brings its share of challenges. Catastrophic forgetting~\cite{forgetting, ewc} is one of the challenges that has been addressed the most in the literature. In task-IL, it is described as the drop in performance on previous tasks when the learning of a new task occurs. While catastrophic forgetting can be tough to solve in some settings and without access to old samples, when a memory buffer of reasonable size is used (as it is done in class-incremental learning replay methods), it becomes easier to avoid and many works have shown how to prevent it \cite{gem, agem}. However, other challenges have been identified, among them, the one of knowledge transfer between tasks. Primarily discussed in the multi-task learning literature \cite{multitask}, the transfer of knowledge between tasks enables the learner to boost its performance on separate tasks when it is learning them concurrently. In the task-incremental learning setting, this transfer between tasks has been referred to as backward- and forward- transfer in GEM~\cite{gem}. Where backward denotes the positive influence on previous tasks performance that learning subsequent tasks could have, while forward denotes the positive influence learning previous tasks could have on the learning of new tasks.

\subsection{On forgetting in class-incremental learning}
\label{sec:metrics}
Catastrophic forgetting~\cite{forgetting, ewc} has been widely cited as being the main cause of performance drop in many continual learning settings. From task-IL works with two tasks similar to transfer learning~\cite{ewc}, the measurement of forgetting was defined as how much of the performance on the source task decreased when learning the target task. Initially, this measurement was very useful even when moving from two tasks to a sequence of them since each task was evaluated separately. It has been widely observed that training a neural network on a sequence of tasks without accessing data from previous tasks results in catastrophic forgetting of previously learned ones. 
Similar variations of that metric have also been proposed for both task-IL and class-IL~\cite{rwalk, lee2019overcoming, serra2018overcoming}. However, we argue that they do not directly transfer to class-IL since, in that setting, the average accuracy is reported on an ever-more difficult problem, which is classifying between all $C^{n}$ classes. For this reason, looking at the classic forgetting measure in class-IL it is not clear if it captures only the forgetting of previous tasks but also the increasing difficulty of the cumulative task.

\minisection{Task-IL metrics:} Consider that  $a^t_k\!\in\![0, 1]$ denotes the accuracy of task $k$ after learning task $t$ ($k\!\leq\!t$), then the \emph{average accuracy} at task $t$ is defined as $A_t = \frac{1}{t} \sum_{i=1}^{t} a^t_i$. \emph{Forgetting} estimates how much the model forgot about previously learned task $k$ at current task $t$ and is defined as $f^t_k=\max_{i \in \{k,\ldots,t-1\}}  a^i_k - a^t_k$. As with accuracy, this measure can be averaged over all tasks learned so far: \mbox{$F^{t}\!=\!\frac{1}{t-1} \sum_{i=1}^{t-1} f^{t}_{i}$}~\cite{agem}. We denote this version of forgetting as \emph{classic forgetting}.

\tablecifargrowmore{Average accuracy after learning all tasks for CIFAR-100 on ResNet-32 from scratch.}

\minisection{Class-IL metrics:} To consider these metrics for the case of class-IL, we have to replace the notion of separate tasks that is used in task-IL by the cumulative task that is composed of the concatenation of all tasks seen so far $\mathcal{T}_{\Sigma}^k = (C_{\Sigma}^k, D_{\Sigma}^k)$. When evaluating a network on the cumulative task $k$, we propose the following method to predict the label:
\begin{equation}
    \hat{y}_{k}(x; \theta) = \argmax_{c\in C_{\Sigma}^k} f(x; \theta)_c\, ,
\end{equation}
which restricts the output to the $C_{\Sigma}^k$ classes. Note that the predictions can change when you chose another task $j\!\neq\!k$, which is desirable since it makes the task easier when $k$ gets smaller, getting rid of the confusion between classic forgetting and the growing task difficulty discussed above. We further introduce the term \emph{cumulative accuracy} of task $k$ as:
\begin{equation}
    b^{t}_{k} = \frac{1}{|D_{\Sigma}^{k}|}\sum_{x, y \in D_{\Sigma}^{k}}\mathrm{1}_{\{y\}}\!\left( \hat{y}_{k}(x; \theta^{t})\right)\, ,
    \label{eq:cumulative_accuracy}
\end{equation}
which refers to the accuracy obtained on the cumulative task $k$ by the model $\theta^t$, and where $\mathrm{1}_{\{y\}}$ is the indicator function which is $1$ when the prediction is correct, and $0$ otherwise. To summarise the performance of the continual learning approach after $t$ tasks have been learned we simply report $b^{t}_{t}$, which is equivalent to the average accuracy classically reported in class-IL works. We then can analogously define \emph{cumulative forgetting} about a previous cumulative task $k$ at current task $t$ as:
\begin{equation}
    f^{t}_{k}=\max_{i \in \{1,\ldots,t - 1\}}  b^{i}_{k} - b^{t}_{k}\, .
\end{equation}
This measure can be averaged over all tasks learned so far: $F^{t}_{\Sigma}\!=\!\frac{1}{t-1} \sum_{i=1}^{t-1} f^{t}_{i}$. The above defined measurements are an adaptation of the ones from~\cite{agem} to the setting of class-incremental learning, the main difference is that ours consider the incremental cumulative tasks instead of the separate ones.

\figaccjoint{20 25 20 55}{0.49}{Average accuracies on CIFAR-100 splitted in 10 tasks (left) and 20 tasks (right) for $FT^{BAL}_{\text{+}CTF}$ and $FT^{BAL}_{\text{-}CTF}$ using 20 exemplars per class compared to their respective upper-bounds (500 exemplars per class). Mean and standard deviation over 10 runs are reported.}

\section{Experimental Results}
Our implementation is based on the FACIL framework~\cite{class_incr_survey}. We use the same data processing and scenario configurations. We extend it with our proposed baselines to compare with its implementation of state-of-the-art methods. More details are available in the supplementary material. Our code is available at \url{https://github.com/AlbinSou/cross_task_cil}.

\minisection{Datasets:} CIFAR-100~\cite{cifar} contains 60k rgb images of size 32x32x3 divided in 100 classes, each having 500 training images and 100 test images. Since there is no defined validation set, we set apart 10\% of the train set as validation, and keep it the same for all experiments. This is a common setting in class-IL~\cite{il2m, end2end, LUCIR, icarl, bic}. The classes are divided between 10 or 20 task splits to be learned incrementally, with a memory buffer that contains 20 exemplars per class for rehearsal (2,000 total). Exemplars are selected using the herding strategy, which has been shown to be slightly more robust than other strategies~\cite{class_incr_survey}. Data augmentation is applied via padding, random crop and random horizontal flips during training. Finally, task and class orderings~\cite{masana2020class}, are fixed to the commonly used iCaRL seed~\cite{class_incr_survey, icarl, bic}.

Imagenet~\cite{Imagenet} contains 1,000 object classes with different number of samples per class. In our experiments, we use a reduced version composed of its 100 first classes, Imagenet-Subset (as in~\cite{icarl}). Data is pre-processed using 224$\times$224 random crop, normalization and random horizontal flip. We split this dataset into 25 tasks with a random class order fixed for all experiments. Exemplars are handled in the same way as the CIFAR-100 experiments.

\minisection{Network:} we use the commonly paired network for CIFAR-100 experiments: ResNet-32~\cite{resnet}, learned with Stochastic Gradient Descent with a patience scheme. More details about the hyper-parameters used during training can be found in the supplementary material. For Imagenet-Subset, we use ResNet-18, which allows for larger input size and provides more capacity.

\minisection{Upper bounds:} for each of the two baselines that use a limited amount of memory, we consider their respective upper bounds that have access to all data from previous tasks, namely $FT^{BAL}_{\text{+}CTF} (max)$ and $FT^{BAL}_{\text{-}CTF} (max)$.

\figgaps{30 30 30 30}{0.9}{Final average accuracy obtained by each method and their respective upper bounds on CIFAR-100 splitted in 10 tasks (Left) and 20 tasks (Right). The part in red coined "others" can be obtained with better intra-task features, while the orange part is the additional gain obtained when learning cross-task features}

\subsection{Importance of cross-task features}
Results for CIFAR-100 are shown in Table~\ref{tab:cifar100}. For the 10 tasks split, using $FT^{BAL}_{\text{+}CTF}$ performs better than $FT^{BAL}_{\text{-}CTF}$ when using 10 or more exemplars per class, indicating that it is able to learn some cross-task features. For very low number of exemplars, we observe that $FT^{BAL}_{\text{+}CTF}$ is performing poorly compared to $FT^{BAL}_{\text{-}CTF}$. The average accuracy gap after 10 tasks between the two methods is around $\sim\!5\%$ when using 20 exemplars per class, which is the most commonly considered memory size for class-IL. As we move to the 20 tasks split, both methods suffer from a $\sim\!5\%$ drop in performance. Through the comparison of these two losses with their respective upper bounds using the maximum of memory (see Fig. \ref{fig:acc_joint}), we observe that the gap due to not learning cross-task features is of $\sim\!10\%$, and jumps to $\sim\!15\%$ when moving to 20 tasks. This gap is filled in part ($\sim\!5\%$ for 10 tasks and $\sim\!7\%$ for 20 tasks) by $FT^{BAL}_{\text{+}CTF}$ when 20 growing memory per class is used. It means that the gap due to the learning of cross-task features is already filled by a half by $FT^{BAL}_{\text{+}CTF}$, the remaining performance gap is then due to something else, materialised by the difference between the upper bound for $FT^{BAL}_{\text{-}CTF} \ (max)$ and $FT^{BAL}_{\text{-}CTF} \ (20)$. We summarise these observations in Fig.~\ref{fig:gaps}.

On Imagenet-Subset (see Fig.~\ref{fig:imagenet}), we observe the same conclusions. There is a consistent difference between the two baselines due to the additional learning of cross-task features. But this difference does not account for the main part of the performance gap, which is observed when comparing $FT^{BAL}_{\text{-}CTF} \ (max)$ and $FT^{BAL}_{\text{-}CTF} \ (20)$, and is not related to cross-task features.

\tableforgetting{Average forgetting (classic) compared to newly defined cumulative forgetting measure on CIFAR100 10 and 20 tasks split for $FT^{BAL}_{\text{+}CTF}$. Cumulative forgetting shows no sign of performance drop due to a forgetting phenomenon.}

\figimnet{20 25 20 55}{1.05}{Average accuracies on Imagenet-Subset splitted in 25 tasks for $FT^{BAL}_{\text{+}CTF}$ and $FT^{BAL}_{\text{-}CTF}$ using 20 exemplars per class compared to their respective upper-bounds with maximum memory. As it is the case on cifar, the gap due to the learning of cross-task features is not predominant, and is partly filled by the use of $FT^{BAL}_{\text{+}CTF}$.}

\subsection{Cumulative accuracy and cumulative forgetting}
We have argued that the forgetting measure defined for task-IL cannot directly be applied to class-IL (as done in some papers). Instead the definitions should be adapted to consider cumulative accuracy of the tasks (see Eq.~\ref{eq:cumulative_accuracy}).

We apply the metrics defined in Sec.~\ref{sec:metrics} to the experiments on CIFAR-100 (10 and 20 tasks). One insightful way to analyse the learning process of class-IL methods is provided in Fig.~\ref{fig:cumulative_accs}, we coin this a \emph{cumulative accuracy graph}. It allows you to analyse the behaviour of the various cumulative tasks while the learning progresses. For example, the line starting from Task 2 indicates $b_2^t$. There, we can observe an initial positive transfer when learning task 3 and a subsequent planar behaviour where performance remains constant. The important fact to observe from this graph is that while the average accuracy decreases considerably over the course of training, the values of the cumulative accuracy $b_{k}^{t}$ when fixing $k$ are quite stable. The resulting cumulative forgetting measure is consequently very low, and only grows for the smaller memory sizes (see Table~\ref{tab:forgetting}). This means that cumulative forgetting cannot explain on its own the decrease in average accuracy. This strongly contrasts with the values obtained using classic forgetting, which are higher and indicate that most of the performance drop is caused by forgetting. In conclusion, we argue that the classic forgetting method is inapplicable to class-IL and instead encourage the usage of cumulative forgetting. More cumulative accuracies graph for other class-IL methods are made available in the supplementary.

Finally, for the model using maximum memory (see Fig.~\ref{fig:cumulative_accs}), we observe that the cumulative accuracies grow over the course of training, which means the model gets better at previous cumulative tasks. This is showing that positive backward transfer is occurring: learning new tasks improves performance on older tasks.

\figaccprognew{12 25 40 60}{0.49}{12 25 40 60}{Cumulative accuracies $b_{k}^t$ on CIFAR100 (10 tasks) for $FT^{BAL}_{\text{+}CTF} (20 mem/cls)$ (Left) and $FT^{BAL}_{\text{+}CTF} (max)$ (Right). Grey dashed lines represent $b_{k}^t$ for varying $t$ and one fixed $k$ per line. The blue dotted line represent $b_{t}^t$ for varying $t$, which is the average accuracy.}
\figcomparison{20 25 20 55}{1.05}{Comparison including multiple class incremental learning methods and the two baselines we use. CIFAR100 10 tasks, 20 growing memory per class. The baselines used perform comparatively to other state of the art methods.}

\subsection{Comparison to state-of-the-art}
In Fig.~\ref{fig:soa_comparison}, we display a comparison of the considered baselines to other state of the art methods like iCarl~\cite{icarl}, LUCIR~\cite{LUCIR}, BiC~\cite{bic} and EEIL~\cite{end2end}. \textbf{FT-E} is another baseline which is similar to $FT^{BAL}_{\text{+}CTF}$ but does not use a second fine-tuning stage. We observe that $FT^{BAL}_{\text{+}CTF}$ and $FT^{BAL}_{\text{-}CTF}$ perform comparatively to the latter methods in this setting, the comparison with \textbf{FT-E} shows the gain of the additional fine-tuning stage, which helps to get rid of the task-recency bias. The main purpose to include this comparison to state-of-the-art methods is to show that the baseline $FT^{BAL}_{\text{+}CTF}$ gets competitive results which means that our analyses on the challenges for $FT^{BAL}_{\text{+}CTF}$ could generalize to other state-of-the-art methods.

\subsection{Discussion}
We here explore the efficiency of cross-task features learning done by $FT^{BAL}_{\text{+}CTF}$ as a function of the memory size. 
To do so, we use two additional metrics that focus on specific points of the learning. The task-aware accuracy $\mathcal{A}_{taw}$ is the accuracy of the model on the test data when it is provided with the task-id, averaged over all tasks the model has seen so far. It is commonly used in task-IL and aims to evaluate the task-specific performance across tasks. Since class-incremental learning comes with the additional challenge of discriminating across tasks, we additionally use a task-inference accuracy $\mathcal{A}_{tinf}$. The latter is computed by counting the number of predictions that fall into the correct task. We report the values of these two measurements with an increasing memory size in Fig. \ref{fig:task_inf}.

Although the use of $FT^{BAL}_{\text{+}CTF}$, explicitly learning cross-task features accounts for a part of the gap between 20 memory and the maximum amount, we observe that both task-aware accuracy and task inference accuracy grow when increasing the memory size. Since forgetting is eliminated as a possible cause (see Tab.~\ref{tab:forgetting}), the improvement in task-aware accuracy when the memory increases is most likely explained by knowledge transfer between tasks. This happens for both $FT^{BAL}_{\text{+}CTF}$ and $FT^{BAL}_{\text{-}CTF}$ and indicates that there is a greater potential gain in task-inference by learning better quality task-specific features (For instance, through improved backward and forward transfer) rather than by learning better cross-task features. Indeed in this case, the learning of cross-task features is already maximum at the 50 exemplars mark (the gap between the red curve and blue curve does not increase anymore).

\figtaskinf{20 25 20 60}{1.0}{Task inference and Task aware accuracy obtained at the end of the task sequence for different memory sizes. CIFAR100 10 tasks. The gap in task inference between $FT^{BAL}_{\text{+}CTF}$ and $FT^{BAL}_{\text{-}CTF}$ stabilises after 50 exemplars per class. Additional performance gains can be obtained by increasing task-aware accuracy, which is correlated with task inference accuracy.}

\section{Conclusion and future directions}
\label{conclusion}
Through the ablation of one major component of class-incremental learning that makes it different from task-incremental learning, we attempted to link the challenges met by these two settings. By studying the impact that the absence of explicit cross-task features learning could have on popular benchmarks, we have shown that the following is already partly solved by the use of a simple replay procedure. While the learning of cross-task feature could still be improved, its influence on the final result may not be as decisive as we might think. Instead, the major part of the gap could be due to other sources. The lack of knowledge transfer studied in task-IL could be one of them. We have also highlighted that forgetting as defined for task-IL has to be carefully adapted for class-IL. Using our proposed cumulative forgetting measure, we observed that forgetting is not the main cause of performance drop in that setting.

We hope that future research can draw more links between task-IL and class-IL. While these two settings differ, we saw that they share similar challenges. In particular, we think that it could be interesting to tackle the task-IL setting using similar memory constraints as in class-IL, and aim for better knowledge transfer instead of zero forgetting. Indeed, enabling more knowledge transfer between tasks could directly be applied to class-incremental learning by learning classification heads on top of the learned feature extractor, similarly to what we did with $FT^{BAL}_{\text{-}CTF}$).

Another promising type of approaches to improve the quality of the learnt representation is learning with an unlabelled data stream concurrently to the current task, as done in~\cite{lee2019overcoming}. That way knowledge is transferred between the data stream and the current task instead of in-between tasks. However, this requires to have access to such a data stream with nice properties w.r.t the tasks at hand.

\section*{Acknowledgments}
We acknowledge the support from Huawei Kirin Solution, the Spanish  project  PID2019-104174GB-I00 (MINECO, Spain), and the CERCA Programme of Generalitat de Catalunya.

\bibliography{references}
\bibliographystyle{icml2021}

\clearpage

\appendix

\section{Hyperparameter selection for CIFAR-100}
We use Stochastic Gradient Descent (SGD) with momentum as the optimization algorithm, along with a learning rate scheduling strategy using patience: the learning rate decreases when the validation loss of the current task does not decrease for a defined number of epochs, the model that performs best on validation data is retained. This patience scheme is also used during the second (fine-tuning) step. Weight decay is also used for regularisation.

The proposed baselines only have an hyperparameter, which is the number of balancing finetuning epochs from step 2. We perform GridSearch/CHF following the procedure in~\cite{delange2021continual, class_incr_survey} that respects the assumptions of continual learning (cannot access future tasks validation data). We perform a learning rate search by fine-tuning on the new task (without applying any other loss or strategy). Once the best learning rate is found for that task we set it as the starting learning rate for it and train the final version of the current task before moving onto the next. Below are listed the parameters used by the grid search.

\begin{verbatim}
    lr_first: [5e-1, 1e-1, 5e-2],
    lr: [1e-1, 5e-2, 1e-2, 5e-3, 1e-3],
    lr_searches: [3],
    lr_min: 1e-4,
    lr_factor: 3,
    lr_patience: 10,
    clipping: 10000,
    momentum: 0.9,
    wd: 0.0002
\end{verbatim}

\section{Additional results}

\subsection{Linear probing of intermediate layers}

\begin{figure}
    \centering
    \begin{subfigure}[b]{\linewidth}
        \includegraphics[width=1.0\linewidth]{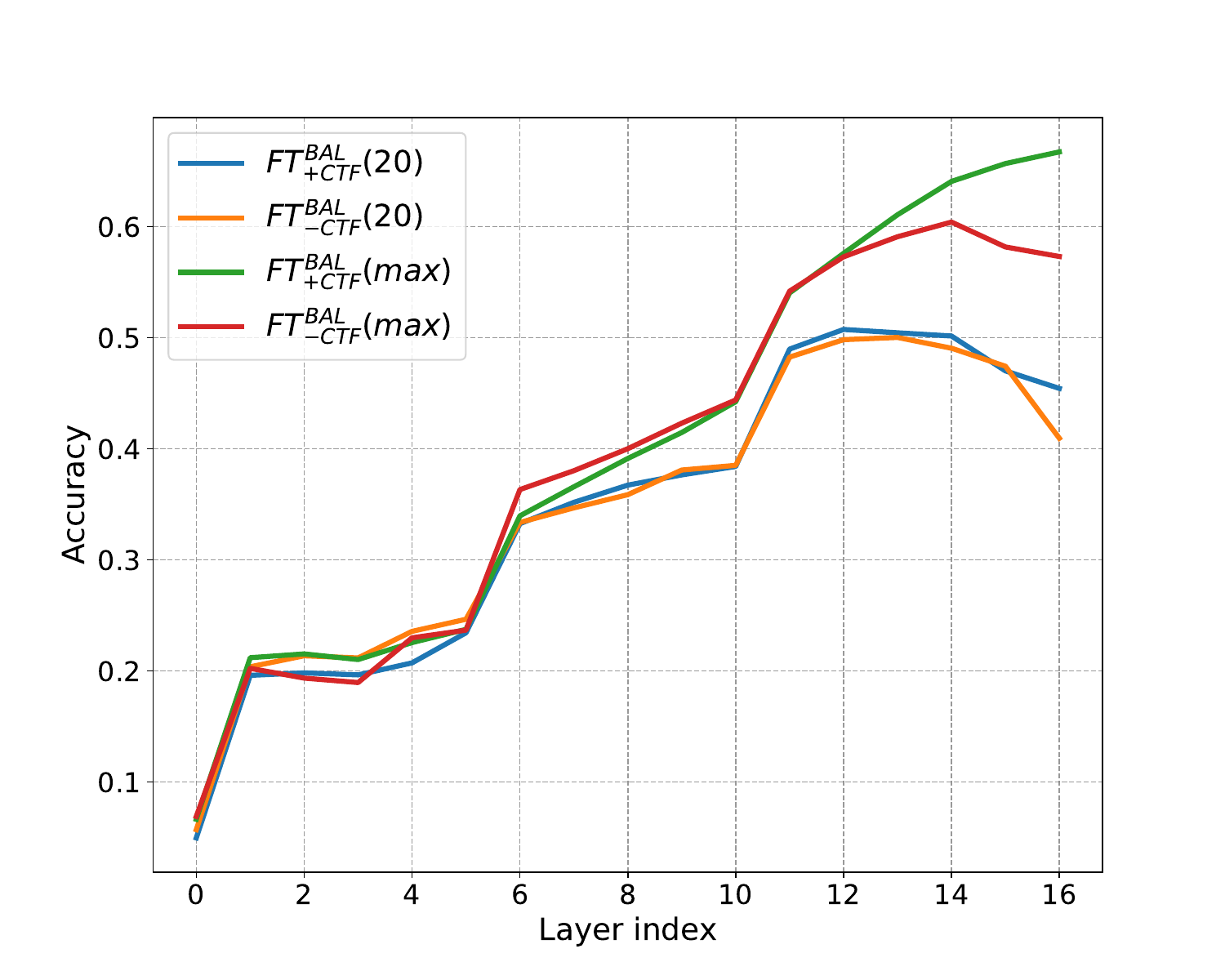}
    \end{subfigure}
    \begin{subfigure}[b]{\linewidth}
        \includegraphics[width=1.0\linewidth]{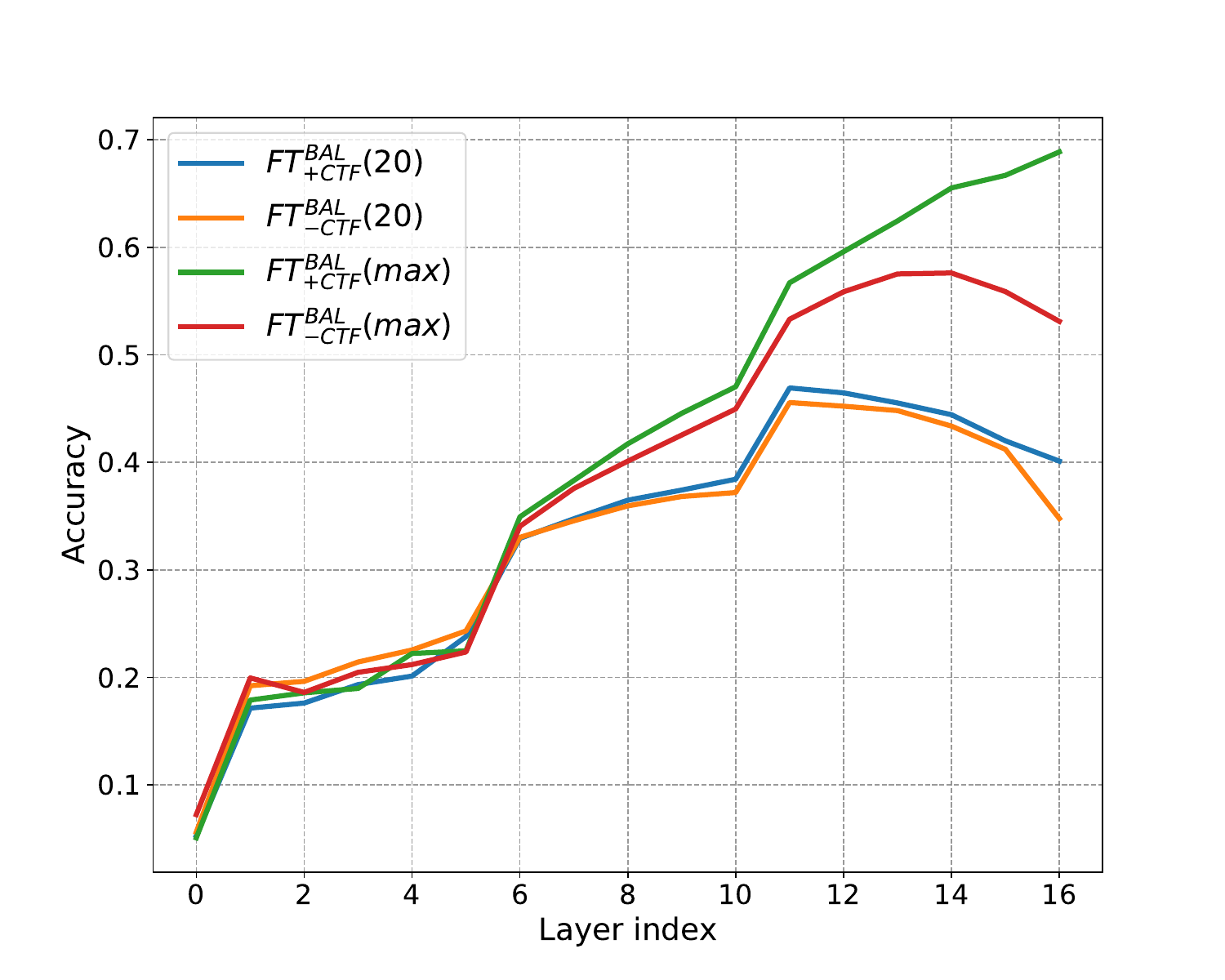}
    \end{subfigure}
    \caption{Final average accuracy after linear probing, for $FT^{BAL}_{\text{+}CTF}$ and $FT^{BAL}_{\text{-}CTF}$ using limited amount of memory (20 exemplars per class) as well as their respective upper bound. Linear probes are learned on final model checkpoints after continually training on CIFAR-100 splitted in 10 tasks (Top) and in 20 tasks (Bottom).}
    \label{fig:cifar100_probe}
\end{figure}

In this section, we aim to evaluate the quality of the representations learned by both $FT^{BAL}_{\text{+}CTF}$ and $FT^{BAL}_{\text{-}CTF}$ methods in terms of linear probing accuracy \cite{alain2017understanding, davari2022probing} at different levels in the network. This technique is widely used in unsupervised learning in order to check the quality of a learned representation, and while it cannot be applied in a realistic continual learning setting, because it requires access to all previous data, it is a good way to evaluate learned representations. In Figure ~\ref{fig:cifar100_probe} we plot the accuracy of linear probes learned on top of each two layers output of the resnet32, and we compare the results for both $FT^{BAL}_{\text{+}CTF}$ and $FT^{BAL}_{\text{-}CTF}$ that used a limited amount of memory during training (20 exemplars per class) as well as using all the data (incremental joint).

We first observe that the difference between $FT^{BAL}_{\text{+}CTF}(max)$ and $FT^{BAL}_{\text{-}CTF}(max)$ is not so marked in the early layers, and more marked in the later layers (Starting from layer index 10 in the figure). We also see that this difference is created at earlier layers and is more important for checkpoints trained on CIFAR-100 splitted in 20 tasks, which makes sense since cross-task features are more important the more tasks they are. Another interesting observation is that while for $FT^{BAL}_{\text{+}CTF}(max)$ the maximum accuracy is reached at the last layers, this is not the case for $FT^{BAL}_{\text{-}CTF}(max)$, where earlier layers from the last block have a better probed accuracy. This observation is consistent with the ones from \cite{sariyildiz2023no}, that claim that representations that are learned in a supervised manner are not optimal for transfer to other tasks when taken at the last layer. In this specific case, the representation learned by $FT^{BAL}_{\text{-}CTF}(max)$ is optimal for the last task of the stream (and all other tasks considering their task-specific classes separately), but not for the full task considering all classes at once. 

Interestingly, for the continual methods that have learned with reduced memory, we see that in both cases (10 and 20 tasks), the features learned by the continual learning methods $FT^{BAL}_{\text{+}CTF}(20)$ and $FT^{BAL}_{\text{-}CTF}(20)$ do not differ a lot until the very last few layers. In general, both of these curves are very close until the very end and the main separation happens after the average pooling layer, both of them have a shape that is more similar to the one of $FT^{BAL}_{\text{-}CTF}(max)$ than to the one of $FT^{BAL}_{\text{+}CTF}(max)$ (with the best accuracy occurring at a layer that is not the last one). This shows that the network really struggles even with 20 exemplars per class in memory to learn any cross-task features in the earlier layers, but is only able to learn some cross-task features in the very last layers.

\subsection{Fixed memory size}
\begin{figure}
    \centering
    \begin{subfigure}[b]{\linewidth}
        \includegraphics[width=1.0\linewidth]{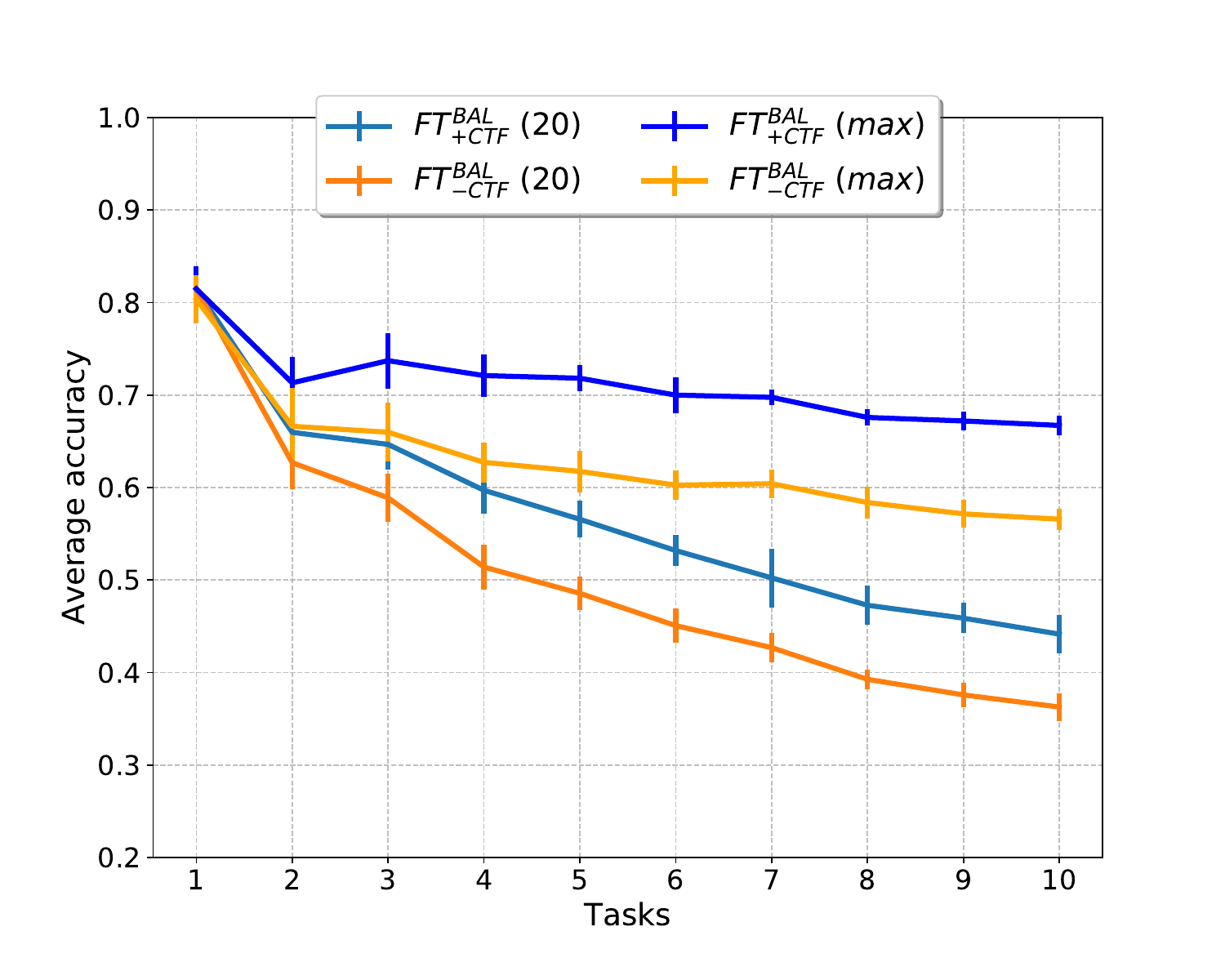}
    \end{subfigure}
    \begin{subfigure}[b]{\linewidth}
        \includegraphics[width=1.0\linewidth]{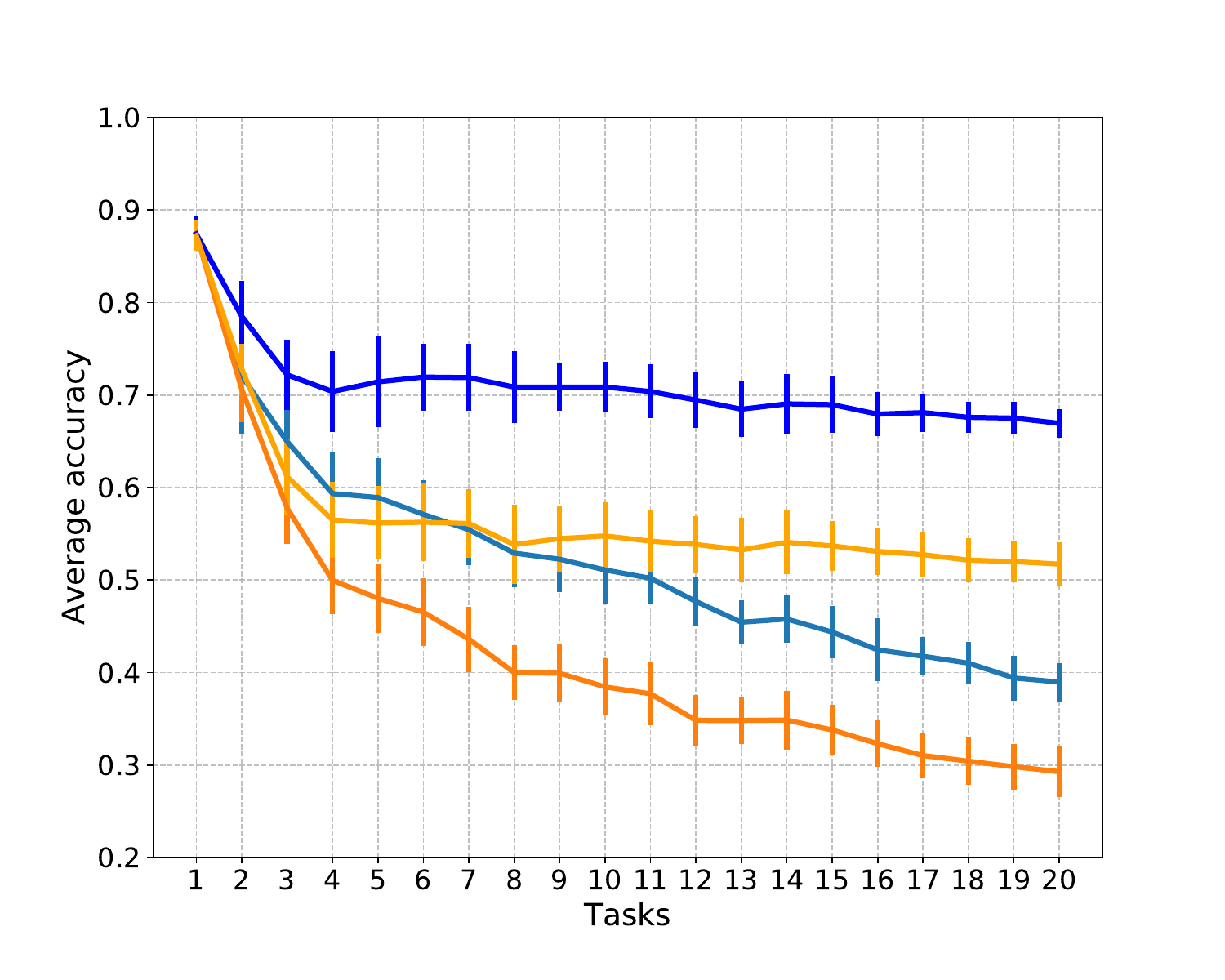}
    \end{subfigure}
    \caption{Average accuracies on CIFAR-100 splitted in 10 tasks (top) and 20 tasks (bottom) for $FT^{BAL}_{\text{+}CTF}$ and $FT^{BAL}_{\text{-}CTF}$ using a fixed memory of 2000 exemplars compared to their respective upper bounds. Mean and standard deviation over 10 runs are reported.}
    \label{fig:fixd}
\end{figure}

In the main article we considered a growing memory size. Results with fixed memory size are higher but we found them harder to interpret since the number of samples per class available varies over training, which makes the incremental learning problem easy in the first few tasks and then gradually harder. Nevertheless, we report results using fixed memory sizes below (see Fig. \ref{fig:fixd}). In this case for the first few tasks $FT^{BAL}_{\text{+}CTF}$ is able to outperform the upper bound that does not learn cross-task features, this is because enough memory is available at that time. For instance, on CIFAR100 (20 tasks) the gap between $FT^{BAL}_{\text{+}CTF} (2000)$ and $FT^{BAL}_{\text{-}CTF} (2000)$ is the same than the gap between $FT^{BAL}_{\text{+}CTF} (max)$ and $FT^{BAL}_{\text{-}CTF} (max)$ at task 3, hinting that $FT^{BAL}_{\text{+}CTF} (2000)$ was able to learn correct cross-task features until that point, as new tasks come however, less memory per class is available, rendering the learning of the latter harder.

\subsection{Cumulative accuracy}
\begin{figure}
    \centering
    \begin{subfigure}[b]{\linewidth}
        \includegraphics[width=1.0\linewidth]{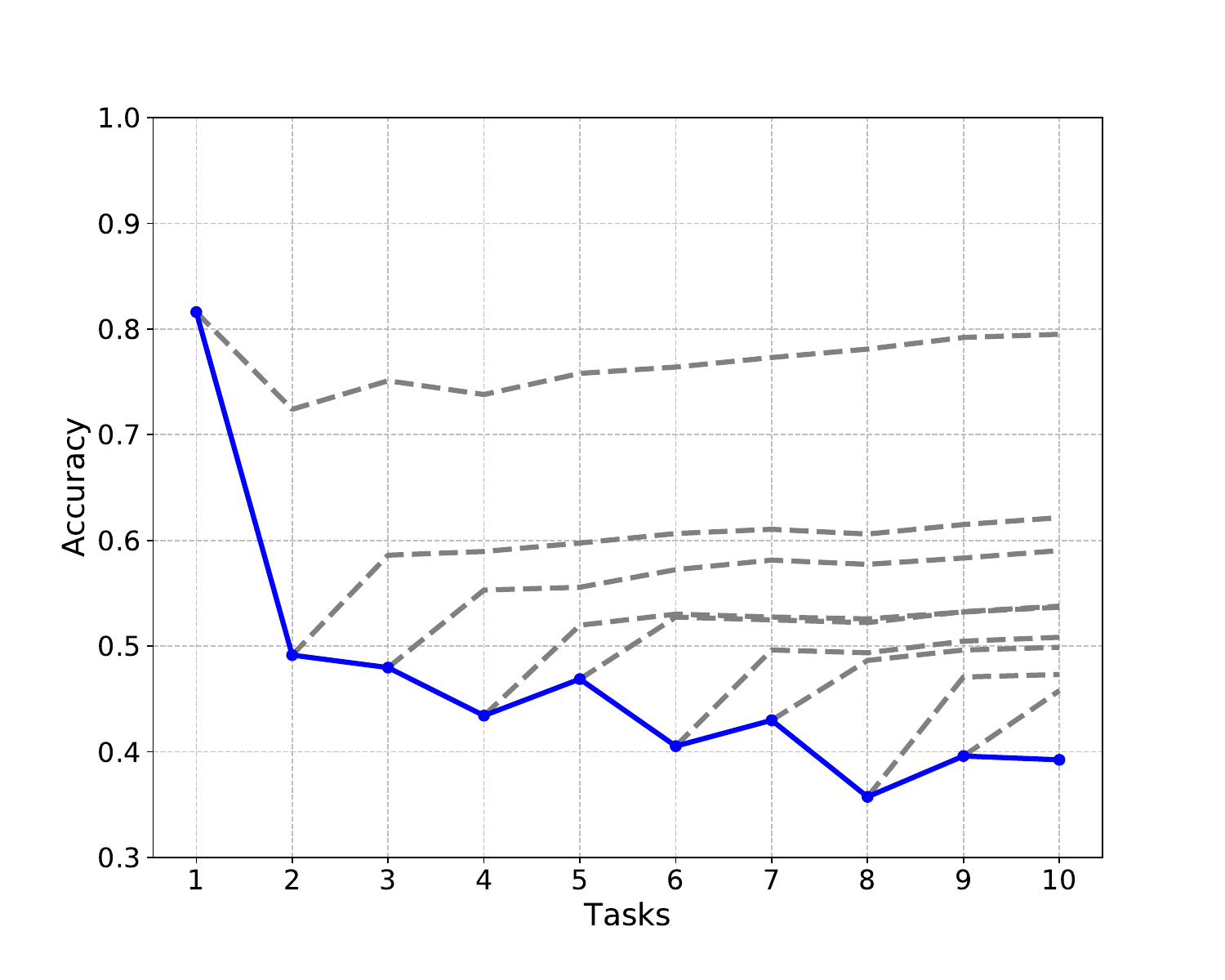}
        \caption{End to End incremental  learning}
    \end{subfigure}
    \begin{subfigure}[b]{\linewidth}
        \includegraphics[width=1.0\linewidth]{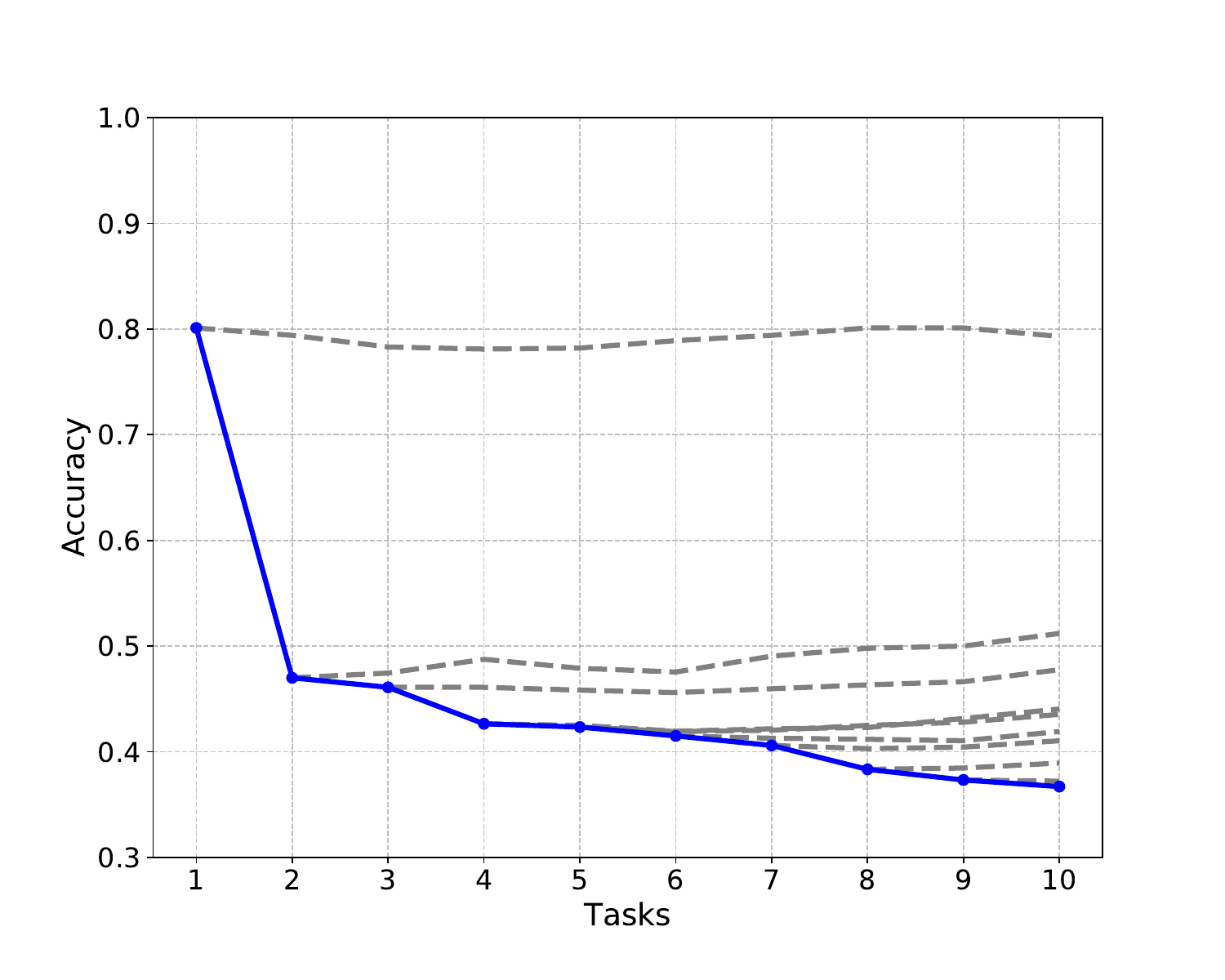}
        \caption{BiC}
    \end{subfigure}
    \caption{Cumulative accuracies $b_{k}^t$ on CIFAR100 (10 tasks) for EEIL (Top) and Bic (Bottom), both using a growing memory of 20 exemplars per class. Grey dashed lines represent $b_{k}^t$ for varying $t$ and one fixed $k$ per line. The blue dotted line represent $b_{t}^t$ for varying $t$, which is the average accuracy.}
    \label{fig:cumulative_others}
\end{figure}

We applied our cumulative accuracy metric to two other state of the art methods, BiC~\cite{bic} and EEIL~\cite{end2end}, results are reported in Fig.~\ref{fig:cumulative_others}. We observe similar results than the ones reported for our baselines. EEIL has a characteristic jump in cumulative accuracy after each task, just like our baselines. Since this method also uses a fine-tuning step we believe this jump is due to the latter. On BiC, the cumulative accuracies are very stable over the course of training.

\begin{table}
\setlength\tabcolsep{4pt}
\centering
\caption{Final average accuracy obtained by both baselines on Imagenet-subset (25 tasks)}
\begin{tabular}{c@{\hspace{0.5cm}}c@{\hspace{0.5cm}}cc}
\toprule
 memory & 20/cls & max\\
\midrule
  $FT^{BAL}_{\text{+}CTF}$  & 31.7 & 70.6 \\
 $FT^{BAL}_{\text{-}CTF}$ & 26.4 & 62.1 \\
\bottomrule
\end{tabular}
\label{tab:imagenet}
\end{table}

\end{document}